\documentclass[lettersize,journal]{IEEEtran}
\usepackage{amsmath,amsfonts}
\usepackage{algorithmic}
\usepackage{array}
\usepackage[caption=false,font=normalsize]{subfig}
\usepackage{textcomp}
\usepackage{stfloats}
\usepackage{url}
\usepackage{verbatim}
\usepackage{graphicx}
\usepackage{booktabs}   
\usepackage{caption}
\usepackage{amssymb}
\usepackage{color}
\usepackage{subfloat} 
\usepackage{hyperref}
\newcommand{\tabincell}[2]{\begin{tabular}{@{}#1@{}}#2\end{tabular}}
\def\BibTeX{{\rm B\kern-.05em{\sc i\kern-.025em b}\kern-.08em
    T\kern-.1667em\lower.7ex\hbox{E}\kern-.125emX}}
\usepackage{balance}
\begin{document}
\title{Self-control: A Better Conditional Mechanism for Masked Autoregressive Model}
\author{Qiaoying Qu, Shiyu Shen
\thanks{Qiaoying Qu and Shiyu Shen are with the School of Statistics and Data Science, KLMDASR, LEBPS, and LPMC, Nankai University, Tianjin 300071, and also with the Key Laboratory of Pure Mathematics and Combinatorics, Ministry of Education, China. (e-mail: quqiaoying@mail.nankai.edu.cn; shenshiyu@mail.nankai.edu.cn).
}}

\markboth{Journal of \LaTeX\ Class Files,~Vol.~18, No.~9, September~2020}%
{How to Use the IEEEtran \LaTeX \ Templates}

\maketitle

\begin{abstract}
Autoregressive conditional image generation algorithms are capable of generating photorealistic images that are consistent with given textual or image conditions, and have great potential for a wide range of applications. Nevertheless, the majority of popular autoregressive image generation methods rely heavily on vector quantization, and the inherent discrete characteristic of codebook presents a considerable challenge to achieving high-quality image generation. To address this limitation, this paper introduces a novel conditional introduction network for continuous masked autoregressive models. The proposed self-control network serves to mitigate the negative impact of vector quantization on the quality of the generated images, while simultaneously enhancing the conditional control during the generation process. In particular, the self-control network is constructed upon a continuous mask autoregressive generative model, which incorporates multimodal conditional information, including text and images, into a unified autoregressive sequence in a serial manner. Through a self-attention mechanism, the network is capable of generating images that are controllable based on specific conditions. The self-control network discards the conventional cross-attention-based conditional fusion mechanism and effectively unifies the conditional and generative information within the same space, thereby facilitating more seamless learning and fusion of multimodal features.

\end{abstract}


\section{Introduction}
Autoregressive model has become a paradigm for sequence modeling due to its robust predictive capabilities. Visual autoregressive models typically flatten an image to a sequence and utilize autoregressive models to predict the subsequent unit. According to the difference of generated sequences, the current mainstream visual autoregressive models can be categorized into three types: pixel-based models, token-based models, and scale-based models.

Pixel-based autoregressive models flatten the image to a sequence of pixels and perform pixel-level predictions. PixelRNN/CNN \cite{Oord2016PixelRN}  models the discrete probability of the raw pixel values and encodes the complete set of dependencies in the image. The Gated PixelCNN \cite{Oord2016ConditionalIG} achieves feature extraction and control at the pixel level by introducing a gating mechanism in the process of generating images to produce high-quality images.

The token-based autoregressive model is inspired by the GPT model in NLP. The image is compressed into discrete tokens by the encoder, which are then employed in an autoregressive model. LlamaGen \cite{Sun2024AutoregressiveMB} applies original “next-token prediction” paradigm of large language models to visual generation
domain. MAGVIT \cite{Yu2022MAGVITMG} introduces a 3D tokenizer to quantize a video into spatial-temporal visual tokens and propose an embedding method for masked video token modeling to facilitate multi-task learning.

The scale-based autoregressive model generates images from coarse to fine using visual token maps of varying scales as autoregressive units. This approach results in the generation of a comprehensive token map for each prediction, which more effectively captures the local structural and detail information of the image. VAR \cite{Tian2024VisualAM}  redefines the autoregressive learning on images as coarse-to-fine “next-scale prediction” or “next-resolution prediction”, diverging from the standard raster-scan "next-token prediction".

However, the sequence discretization operation of the aforementioned image autoregression methods sacrifices the image reconstruction accuracy in order to ensure the property of the codebook. Furthermore, the utilization of the codebook also affects the quality of the image generation. MAR \cite{Li2024AutoregressiveIG} achieves continuous image autoregressive generation by replacing vector quantization with diffusion loss, thus resolving the issue of low generation accuracy. Fluid \cite{Fan2024FluidSA} introduces text conditioning to achieve continuous text-to-image autoregressive generation. 

This paper proposes a novel conditional fusion mechanism that integrates a multimodal conditional generation process into a single self-attention mechanism. In particular, this methodology concatenates multimodal conditions with the images to be generated to obtain a new sequence. This sequence is then subjected to autoregressive generation, thereby providing a more seamless and efficient method for controlled image generation.

The contributions of our algorithm can be summarized as follows.

$\bullet $ We propose a self-control network, which realizes multimodal conditional image generation based on continuous autoregressive model.

$\bullet $ We introduce a conditional mechanism, which unifies multimodal conditions into a single self-attention.

\section{RELATED WORK}
\subsection{Visual Autoregressive Model}
Autoregressive models are typically employed to deal with discrete data, whereas images are continuous therefore need discretization. Vector quantization \cite{Oord2017NeuralDR} became the dominant discretization method for autoregressive visual models \cite{Esser2020TamingTF, Sun2024AutoregressiveMB, Li2024ScalableAI, Yu2022MAGVITMG}, which turns continuous images into discrete images, and also increases the efficiency through reducing the image size. 

Since the autoregressive step-by-step prediction approach is not optimal for images, MaskGIT \cite{Chang2022MaskGITMG} employs a non-autoregressive decoding method, which synthesizes an image in constant number of steps. MAGE \cite{Li2022MAGEMG} unifies generative model and representation learning by a single token based masked image modeling framework with variable masking ratios, introducing new insights to resolve the unification problem.

Furthermore, VQ discretization sacrifices the image reconstruction accuracy due to the property of the codebook. In order to eliminate the effects of discretization, recent studies have indicated that the VQ may not be a prerequisite for autoregressive image generative models. GIVT \cite{Tschannen2023GIVTGI} generates vector sequences with real-valued entries, instead of discrete tokens from a finite vocabulary. MAR \cite{Li2024AutoregressiveIG} is proposed to model the per-token probability distribution using a diffusion procedure, which allows to apply autoregressive models in a continuous-valued space.

\subsection{Text-to-image Generation Model}
Currently, text-to-image generation models primarily categorized into three types: Generative Adversarial Network (GAN) \cite{Goodfellow2014GenerativeAN}, Diffusion Models \cite{Ho2020DenoisingDP}, and Autoregressive Models.

The Generative Adversarial Networks for text-to-image generation comprise three principal components: a text encoder, a generator, and a discriminator. StackGAN \cite{Zhang2016StackGANTT} decomposes the hard problem into more manageable sub-problems through a sketch-refinement process, which addresses the lack of detail in synthetic images. AttnGAN \cite{Xu2017AttnGANFT} allows attention-driven, multi-stage refinement for fine-grained text-to-image generation, which shows that the layered attentional GAN is able to automatically select the condition at the word level for generating different parts of the image. SDGAN \cite{Zhou2022SDGANSD} tackles the significant challenge of textual semantic image generation caused by the bias of linguistic expression variants.

Current the mainstream text-to-image generation algorithms are typically based on diffusion models \cite{Nichol2021GLIDETP, Saharia2022PhotorealisticTD, Ramesh2022HierarchicalTI}. LDM \cite{Rombach2021HighResolutionIS} applies the diffusion and reconstruction process to the latent space, which improves the training and sampling efficiency while guaranteeing the quality of generation. The attention mechanism of transformer has brought significant advances to diffusion model text-to-image generation \cite{Peebles2022ScalableDM, Bao2022AllAW}. In addition, fine-tuning algorithms \cite{Hu2021LoRALA, Ruiz2022DreamBoothFT, Zhang2023AddingCC} have been proposed in order to apply pre-trained weights to specific tasks.

The core concept of autoregressive text-to-image generation techniques is to input image tokens and text tokens as the same sequence into the autoregressive model. The classical autoregressive text-to-image models \cite{Ramesh2021ZeroShotTG, Ding2021CogViewMT} convert text and images to discrete tokens, respectively, and then fed into the autoregressive model, which learns to generate images conditioned on the text. The development of MAR \cite{Li2024AutoregressiveIG} indicates that the VQ may not be a prerequisite for autoregressive image generative models. Thus continuous text-to-image generation model \cite{Fan2024FluidSA} has been proposed. 

\section{PRELIMINARY}
\subsection{Autoregressive Model}
Given a sequence of tokens ${x_1, x_2, ..., x_n}$ where the superscript $1 \leq i \leq n$ specifies an order, autoregressive models formulate the generation problem as “next token prediction”:
\begin{equation}
	p({x^1}, \ldots ,{x^n}) = \prod\limits_{i = 1}^n {p({x^i}\left| {{x^1}, \ldots ,{x^{i - 1}}} \right.)} 
\end{equation}
A network is used to represent the conditional probability $p({x^i}\left| {{x^1}, \ldots ,{x^{i - 1}}} \right.)$.
\subsection{Continuous Autoregressive Model}
MAR \cite{Li2024AutoregressiveIG} rewrites this formulation in two parts: ${z^i} = f({x^1}, \ldots ,{x^{i - 1}})$ and $p({x^i}\left| {{z^i}} \right.)$. Unified with the mask autoregressive model, autoregressive model can be written as:
\begin{equation}
\begin{aligned}
p({x^1}, \ldots ,{x^n}) &= p({X^1}, \ldots ,{X^n})\\ &= \prod\limits_k^K {p(\left. {{X^k}} \right|{X^1}, \ldots ,{X^{k - 1}})} 
\end{aligned}
\end{equation}
where ${X^k} = \{ {x^i},{x^{i + 1}}, \ldots ,{x^j}\} $ is a set of tokens to be predicted at the k-th step, with $\bigcup\nolimits_k {{X^k}}  = \{ {x^1}, \ldots ,{x^n}\} $.
\begin{figure}[!t]
	\renewcommand{\captionfont}
	\centering
	\subfloat[\footnotesize Causal attention ]{
		\includegraphics[width=1.37in]{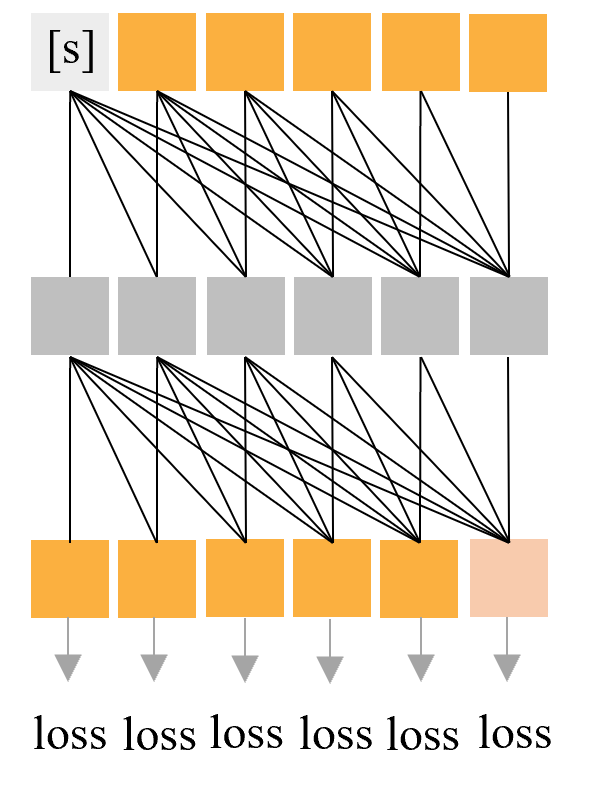}
		\label{causal}}
	\hfil
	\subfloat[\footnotesize Bidirectional attention ]{
		\includegraphics[width=1.37in]{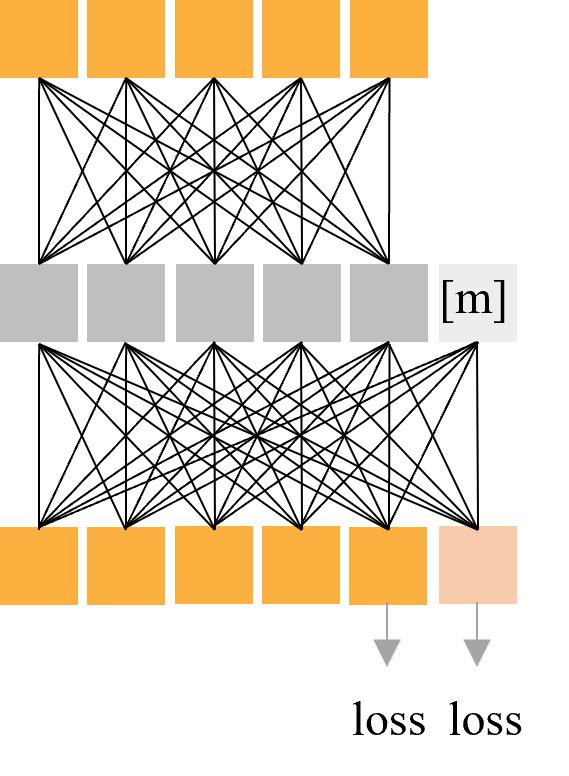}
		\label{bidirectional}}
	\caption{Illustration of the causal attention and bidirectional attention. }
	
\end{figure}
\subsection{Causal Attention $\&$ Bidirectional Attention}
As illustrated in the Fig.\ref{causal}, the causal attention mechanism is a sequential information processing mechanism where the model can only make predictions based on previous information when predicting the current token. This mechanism maintains the coherence and consistency of the generation process and is particularly useful for tasks such as text generation, where it is crucial to maintain the consistency of the temporal structure. In contrast, bidirectional attention, as illustrated in the Fig.\ref{bidirectional}, is an attention mechanism that can take into account both prior and subsequent information, thus enhancing the ability to capture and understand known information. 

\section{METHODOLOGY}
\begin{figure*}[bpht]
	\centering
	\includegraphics[scale=0.70]{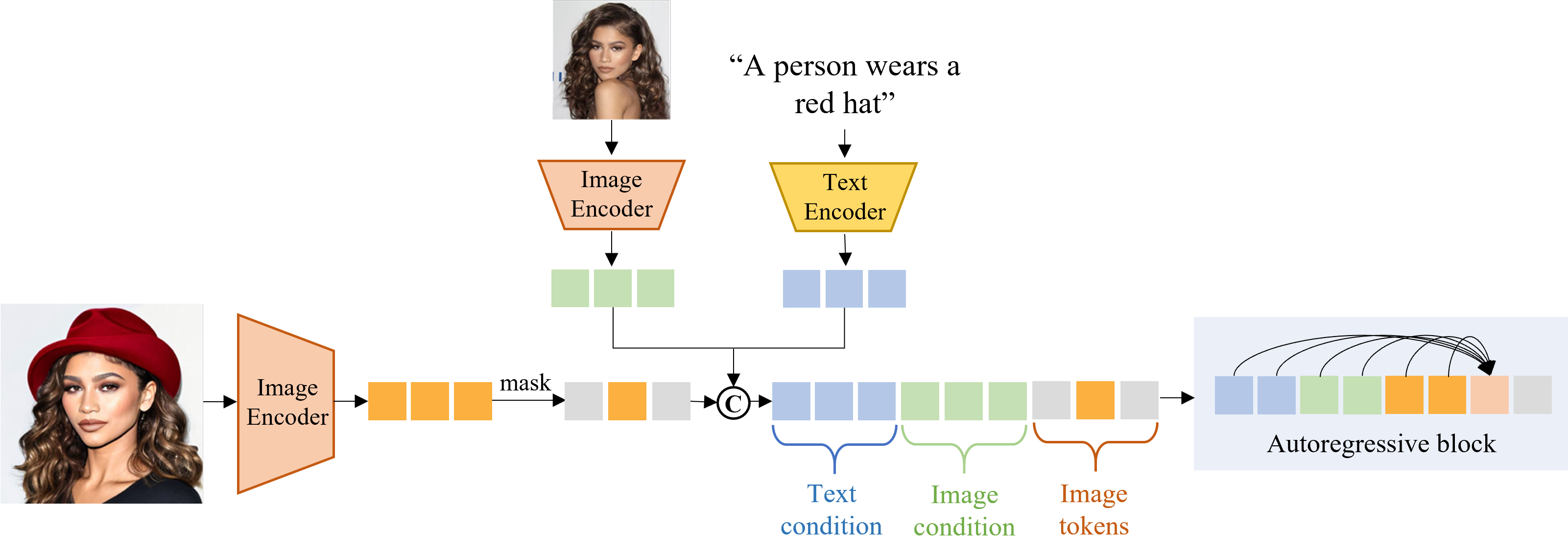}
	\caption{Illustration of self-control network. }
	\label{overview}
\end{figure*}
\begin{figure*}[bpht]
	\centering
	\includegraphics[scale=0.9]{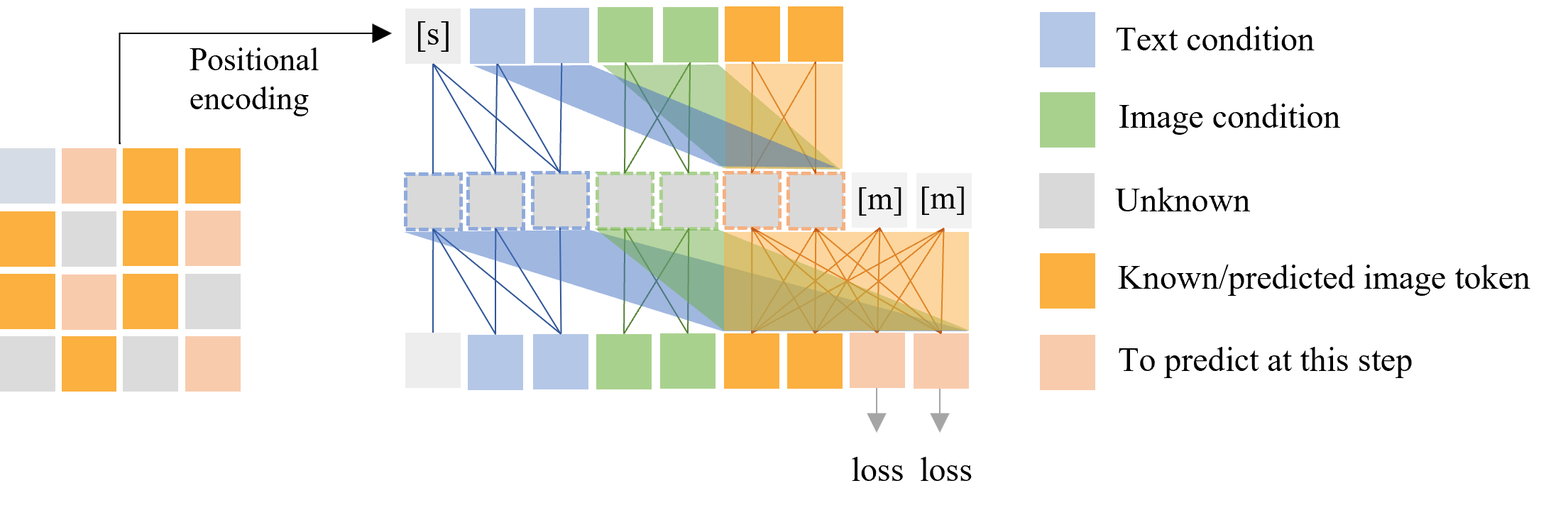}
	\caption{Illustration of attention mechanisms in autoregressive module. }
	\label{attention}
\end{figure*}

\subsection{Conditional Injection}
The network is capable of introducing text and image conditions in order to control the generation of images. As illustrated in the accompanying figure, the condition information from the various modalities is tokenized through the respective text or image encoder, thereby obtaining the corresponding tokens. The generated image tokens are concatenated with the tokens of the two modalities to construct a new sequence, which is then passed through masking operations. This sequence is subsequently fed into the autoregressive module, which performs the image generation through the next token prediction. For the autoregressive module, we employed a MAE-style \cite{He2021MaskedAA} encoder-decoder.

As illustrated in the Fig.\ref{overview}, in the autoregressive module, the text and image condition tokens and the known image token are employed jointly to guide the prediction of the subsequent token. This approach merges text and image conditions, uniting them in a single space and thereby facilitating seamless information integration. The network is designed to process information from different modalities in parallel through the self-attention mechanism. The parallel processing allows for the effective integration and utilization of both the semantic information of textual descriptions and the visual information of image features, which greatly improves the efficiency of information processing and facilitates the control of multimodal conditional information for image generation.

\subsection{Causal Attention for Conditional Text Tokens}
In the autoregressive module, in order to guarantee the effective and precise implementation of textual conditioning for the stepwise generation of images, we employ a causal attention mechanism for the extraction of long-distance features within the textual conditioning, as shown by the blue lines in Fig.\ref{attention}. This mechanism guarantees that the autoregressive model strictly adheres to the temporal constraints when processing text sequences. In text conditioning for image generation, image generation is a step-by-step cumulative process. Each step depends on the content of the image and the text conditioning generated in the previous steps. If the model can maintain strict temporal order constraints when processing text conditioning, it can more accurately understand the textual description and generate images that are highly consistent with the content of the text.

\subsection{Bidirectional Attention for Conditional $\&$ Generated Image Tokens}
In regard to image conditional tokens, the bidirectional attention mechanism is employed, as shown by the green lines in Fig.\ref{attention}. It is capable of integrating forward and backward contextual information to identify broader and deeper semantic dependencies. In contrast to textual sequences, images are a spatial data structure and thus lack a clear sequential order. Consequently, when dealing with image condition tokens, it is not feasible to apply the constraint of temporal order. The bidirectional attention mechanism is able to process all the information before and after the current position simultaneously. This global context awareness allows the model to better understand the semantics of the image conditions and combine them with the entire context of the generated image, thereby producing a more desirable image.

We apply a bidirectional attention mechanism not only to image condition tokens, but also to the generated image tokens, as shown by the orange lines in Fig.\ref{attention}. This approach was selected to more effectively capture the distant relationship between the tokens to be predicted and the pre- and post-generated tokens, thereby facilitating more precise reconstruction of the spatial information. The model will modify the attentional weights in accordance with all the tokens generated prior to and subsequent to the current location when processing the internal tokens of the generated image. These attentional weights reflect the strength of the dependencies between different locations, and the model will leverage these weights to integrate information from disparate locations to achieve more accurate predictions.

\subsection{Multimodal Attention for Multimodal Tokens}
In order to facilitate long-range learning between the three modal tokens, namely the text conditional tokens, image conditional tokens and the generated image tokens, we have adopted the causal attention mechanism, as shown by the blue and green quadrilaterals in Fig.\ref{attention}. In this case, the to-be-generated token is able to learn the long-range relationship between different modalities to generate a high-quality image that is consistent with both the textual descriptions and the image contexts. Furthermore, the condition information of the different modalities and the generated images do not affect the reconstruction of the conditions themselves.

\begin{figure}[bpht]
	\centering
	\includegraphics[scale=0.80]{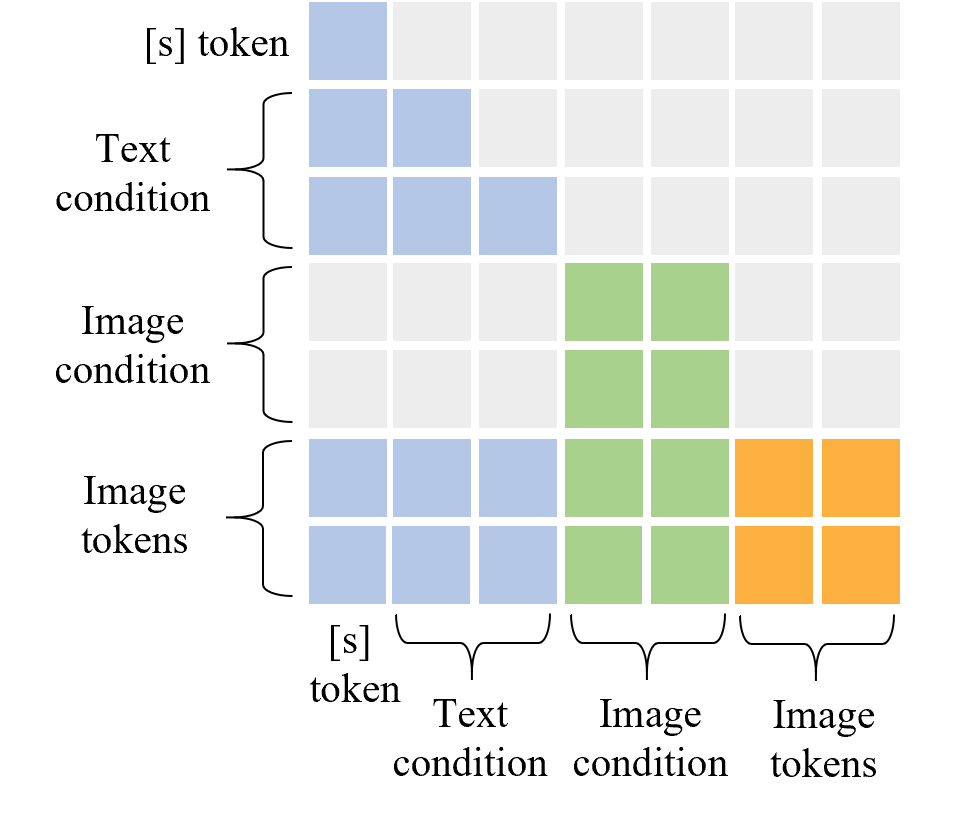}
	\caption{Illustration of the visualization of attention.}
	\label{token}
\end{figure}

\section{EXPERIMENTS}
\subsection{Ablation experiments}

\begin{table*}[ptbh]
	\captionsetup{font={footnotesize}}
	\caption{Quantitative results of ablation experiments on the VEDAI dataset.}
	
	\label{ablation}
	\renewcommand{\arraystretch}{1.5}
	\begin{center}
		\begin{tabular}[c]{c|ccc|cc}
			%[c]{p{1cm}<{\centering}|p{1cm}<{\centering}|p{1.5cm}<{\centering}p{2cm}<{\centering}p{1cm}<{\centering}p{1.5cm}<{\centering}p{2cm}<{\centering}p{1.2cm}<{\centering}}
			\toprule[1pt]\hline
			Options & text & image & multimodal & FID & IS \\\hline
			\tabincell{c}{1\\2\\3\\4\\5\\6\\7\\8\\}
			&\tabincell{c}{ causal \\ causal \\ causal \\ causal \\ bidirectional \\ bidirectional \\ bidirectional \\ bidirectional \\}
			&\tabincell{c}{ causal \\ causal \\ bidirectional \\ bidirectional \\ causal \\ causal \\ bidirectional \\ bidirectional \\}
			&\tabincell{c}{ causal \\ bidirectional \\ causal \\ bidirectional \\ causal \\ bidirectional \\ causal \\ bidirectional \\}
			\\ \hline

			\bottomrule[1pt]
		\end{tabular}
		
		\label{Ablation Study}
	\end{center}
\end{table*}
\small
\bibliographystyle{IEEEtran}
\bibliography{reference}
\end{document}